# Harnessing Large Language Models for Precision Querying and Retrieval-Augmented Knowledge Extraction in Clinical Data Science

## Author Information

Affiliations

**Institute of Psychiatry, Psychology and Neuroscience King's College London, London**

Juan Jose Rubio Jan, Jack Wu

**Institute of Health Informatics, University College London, London**

Julia Ive

## Abstract

This study applies Large Language Models (LLMs) to two foundational Electronic Health Record (EHR) data science tasks: structured data querying (using programmatic languages; Python/Pandas) and information extraction from unstructured clinical text via a Retrieval-Augmented Generation (RAG) pipeline.

We test the ability of LLMs to interact accurately with large structured datasets for analytics and the reliability of LLMs in extracting semantically correct information from free-text health records when supported by RAG.

To this end, we presented a flexible evaluation framework that automatically generates synthetic question–answer pairs tailored to the characteristics of each dataset or task. Experiments were conducted on a curated subset of MIMIC-III, (four structured tables and one clinical note type), using a mix of locally hosted and API-based LLMs.

Evaluation combined exact-match metrics, semantic similarity, and human judgment. Our findings demonstrate the potential of LLMs to support precise querying and accurate information-extraction in clinical workflows.

## Introduction

Electronic Health Records (EHRs) have become the predominant source of patient data in contemporary healthcare, enabled by the digitalisation of clinical workflows and the promise of interoperable decision-support systems (Gunter et al., 2005). In this context, two capabilities are foundational for realising the value of EHRs: structured querying (e.g., SQL, Python/Pandas) to interrogate large relational datasets typical of systems like MIMIC‑III (Johnson et al., 2016), and information-extraction from unstructured clinical notes to identify cohort-defining facts, timelines, and outcomes (Meystre et al., 2008). Large Language Models (LLMs) offer strong advantages on both fronts: they excel at code generation (text‑to‑SQL and Python/Pandas) and at zero‑shot information extraction, which is particularly valuable where annotated data is scarce (Ignashina, M. et al., 2025; Guevara et al, 2024; Alsentzer et al., 2023).

However, LLMs are also capable of "hallucination": producing fluent but unsupported statements (Ji et al., 2023). For EHR applications, where clinical accuracy is of critical importance (safeguarding patient safety and research integrity), rigorous evaluation frameworks are essential to ensure outputs are both factually correct and semantically consistent with source data. In this work, we evaluate the consistency and clinical validity of LLM-generated responses by focusing on two key aspects: (1) assessing whether LLMs can effectively interact with large structured clinical datasets through Python/Pandas to perform precise querying and analytics, and (2) determining how accurately LLMs can extract semantically correct information from free-text health data.

Prior work informs both of our research questions. LLMs have shown strong performance in Text-to-SQL tasks, with prompt engineering techniques such as in-context learning and Chain-of-Thought reasoning significantly improving query generation accuracy (Rajkumar et al., 2022). Recent approaches, such as DIN-SQ and MAG-SQL, further enhance this capability by separating complex questions into smaller steps (Pourreza & Rafiei et al., 2023; Xie et al., 2024). In the context of EHR applications, smaller instruction‑tuned models such as FLAN‑T5 have demonstrated strong performance in off‑the‑shelf evidence detection (e.g., pediatric depression detection or postpartum hemorrhage), surpassing traditional baselines and underscoring their practical utility for cohort selection and analytics pipelines that depend on precise structured queries (Ignashina et al., 2025; Alsentzer et al., 2023). Also, findings from the broader domains indicate that both smaller and larger models can assist in information extraction tasks, with larger models often achieving superior results (Ziems et al., 2023).

Also, to address the hallucination problem in the medical domain, prior work has introduced metrics such as factual consistency scores over QA datasets (Pal et al., 2023). Building on this, Asgari et al. (2025) experiment with various prompting techniques to minimise errors for safer clinical documentation. To further reduce hallucinations, we propose creating domain-specific evaluation datasets by synthetically generating questions with known answers for two key tasks: code generation and information extraction. We follow the approach of Ding et al. (2025), who build silver-standard datasets from NICE guidelines for clinical LLMs.

Our contributions are twofold: (i) a generalisable method for creating test cases for structured and unstructured data applicable to any healthcare use case, and (ii) the application of this framework to the MIMIC-III dataset using a range of LLMs, including locally hosted and API-based models. We demonstrate that this approach enables reliable retrieval and summarisation of clinically relevant information, while also illustrating the practical complexity of deploying such technologies in clinical contexts and assessing the readiness of current models for real-world healthcare applications.

## Datasets and Methods

    a. Datasets

For all tests within the pipeline implementation presented in this work, we used the MIMIC-III database (Johnson et al., 2016), which is a large and freely available resource of deidentified health records. The original dataset contains information on more than forty thousand patients across twenty-six tables. Most of the files provided in the database by Johnson et al. (2016) are distributed in comma-separated values (CSV) format.

For this specific prototype, we focused on a subset of the MIMIC-III database, using only the Patients, Prescriptions, Diagnoses, and D_ICD_Diagnoses tables for the structured data component of this research, and the NoteEvents table for the unstructured data experiments. Additionally, due to the dimensionality of the data and constraints related to our token budget for processing, we restricted the analysis to 101 unique randomly selected patients from the Patients table. The remaining tables were subsequently merged using the SUBJECT_ID and ICD9_CODE fields.

| Category | Description |
|---|---|
| Data source | MIMIC-III (Johnson et al., 2016) |
| Data modality | Structured and unstructured clinical data |
| Structured tables used | PATIENTS, PRESCRIPTIONS, DIAGNOSES_ICD, D_ICD_DIAGNOSES |
| Unstructured table used | NOTEEVENTS |
| Number of patients | 101 |
| Total records | 274,022 |
| Number of features | 23 |
| Join keys | SUBJECT_ID, ICD9_CODE |
| Structured data format | CSV |
| Unstructured data format | Free-text clinical notes |
| Text preprocessing unstructured (RAG) | Chunking (400 tokens, 50-token overlap) |
| Embedding model unstructured (RAG) | MiniLM-L6-v2 |
| Vector index unstructured (RAG) | FAISS |
| Date of birth handling | Synthetic dates generated using Faker |
| Rationale for DOB synthesis | Original dates anonymised and not analytically meaningful |

Table 1. Summary of the MIMIC-III data subset and methodological design

b. Methods
    i. Test data creation

Because all data in MIMIC-III are anonymised, some values in fields such as DOB do not correspond to realistic dates and are therefore not meaningful for analysis. Consequently, we synthetically generated a new field, DOB_Demo, using the faker and datetime libraries, producing random dates within a predefined range. After introducing this field and removing variables that were not required for subsequent analyses and would unnecessarily increase tokenisation costs, the resulting dataset had dimensions of (274,022, 23), corresponding to 274,022 records and 23 features. Although the cohort comprised only 101 unique subject identifiers, both the Prescriptions and Diagnoses tables contain multiple entries per subject. Consequently, joining these tables at the subject level yielded a substantially larger dataset due to the one to many relationships inherent in the underlying clinical data.

For the unstructured data test, we manually selected one clinical note from the NoteEvents table and created a generic text (.txt) file for evaluation. This selection was made because the free-text records in the database vary in length and dimensionality across patients, and because of the previously mentioned limitations related to the number of tokens that could be processed. At the same time, the specific clinical note was chosen based on its completeness and representative nature, as it contained sufficient detail of content to support meaningful testing.

For the structured dataset, we designed 30 prompts of varying complexity, accounting for differences in data preprocessing requirements, aggregation tasks, and the number of operations needed. These factors were treated as shifting variables to enable a comprehensive evaluation of the pipeline. For example, the prompt "What is the median age?" represents a relatively simple two-step operation for the model, as the dataset contains date of birth rather than age, requiring an initial transformation followed by a median aggregation. A follow-up prompt with increased complexity was "What is the median age of female subjects?", which additionally requires conditional filtering prior to aggregation.

For the unstructured dataset, we used a more advanced LLM (GPT-5) to segment the original .txt document into 50 semantically coherent chunks, from which targeted questions were generated. Each question was directly derived from the content of its corresponding text segment. For example, the question "Why was the pre-surgical physical exam not obtained?" was generated from the original sentence: "Physical exam prior to surgery was not obtained since the patient was intubated and sedated." This approach ensured that each question was grounded in the source text and allowed for a systematic evaluation of the model's ability to retrieve and reason over unstructured clinical information.

      ii.      Structured dataset pipeline design

Due to the nature of structured data, we decided to implement a working prototype based on a real-world health informatics practice, using an agent-based pipeline and generating questions in a dynamic way depending on the nature of the data. The size of the dataset and the semantic meaning encoded in its row and column structure make simple Retrieval-Augmented Generation (RAG) approaches and token based ingestion unsuitable for this use case. Instead, we adopted an AI agent capable of translating natural language queries into executable Python (Pandas) code, which is then run within a controlled execution environment. This design allows structured queries to be answered programmatically while preserving the relational semantics of the data, making it a more appropriate approach for interacting with tabular healthcare data.

This approach was implemented for both locally hosted models and proprietary API-based large language models, specifically Llama 3 8B Instruct and GPT-4o Mini, respectively. Llama 3 8B Instruct was selected as one of the smallest viable large language models capable of reliable code generation while remaining suitable for local deployment. This model provided sufficient instruction following and agent capabilities for the structured querying tasks, while maintaining a computational viability with locally hosted experimentation. Llama 3 8B Instruct was accessed via the Hugging Face Transformers interface, enabling local inference and integration within the experimental pipeline for structured data querying and unstructured text generation tasks. The authors of the MIMIC-III repository emphasise the importance of responsible data usage, particularly when interacting with external or online services such as commercial large language model APIs. In line with these recommendations, an additional motivation for the proposed LLM pipeline design was to ensure compliance with approved data handling practices. Specifically, the use of the Azure OpenAI Service was selected because it is listed as an authorised data processor for MIMIC data, thereby allowing the integration of proprietary LLMs while adhering to the repository's data governance requirements.

Both models were provided with the same system prompt to ensure consistent generation of executable code for the CSV agent. In this setup, no component of the tabular data is embedded; instead, assumptions derived from user queries are validated against column names and resolved through executable filtering, aggregation, or transformation operations expressed in Pandas code. Once the pandas code is executed, the LLM preprocess the response and generates an output as illustrated in Figure 1.

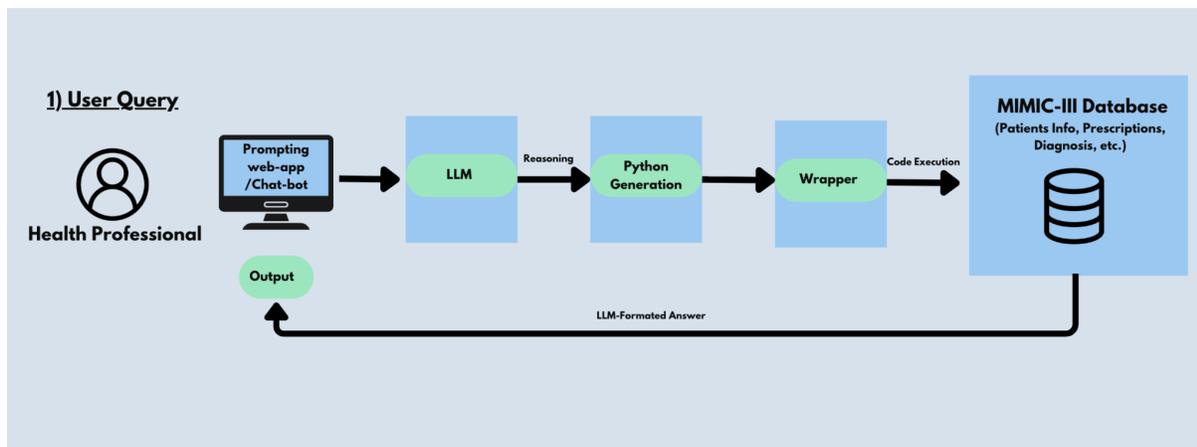

Figure 1. Agent-based csv Pipeline

    iii.    Unstructured dataset pipeline design

As part of a representative pipeline, we implemented a Retrieval-Augmented Generation (RAG) framework, originally introduced by Lewis et al. (2021), to extend the effective knowledge of both locally hosted and proprietary API-based large language models by grounding their outputs in externally retrieved unstructured documents.The models, were specifically Flan-T5 Large and GPT-4o Mini, by providing access to unstructured data. For the unstructured data experiments, FLAN-T5 Large was selected instead of Llama 3 8B Instruct. Despite having a smaller parameter count, FLAN-T5 Large is explicitly trained on a diverse mixture of natural language understanding and instruction-following tasks, making it well suited for text-based question answering and information extraction from free-text clinical notes. In this context, its architecture and training objectives were more aligned with the unstructured extraction task than those of Llama 3 8B, which was prioritised for structured code generation and agent-based interaction. As illustrated in Figure 2, the process begins by uploading the pre-processed and sampled data described in Table 1 into an embedding model, which transforms the documents into vector representations. These embeddings are then stored in a vector database that enables retrieval and interaction with the LLM.

User prompts are similarly pre-processed and embedded using the same embedding model, in this case MiniLM-L6-v2, in order to generate a shared vector space. This allows relevant document chunks to be retrieved and passed to the LLM to support query answering. The vector store was constructed using a consistent chunking strategy, with a chunk size of 400 tokens and an overlap of 50 tokens, and was subsequently indexed using FAISS (Facebook AI Similarity Search), a library for efficient similarity search and clustering of dense vectors (Johnson, Douze, and Jégou, 2017).

The Retrieval-Augmented Generation (RAG) paradigm enables the seamless integration of external data sources with LLMs, allowing users to query domain-specific information while benefiting from the reasoning capabilities of pre-trained models. This approach extends model knowledge without requiring fine-tuning, making it a flexible and computationally efficient strategy for incorporating external unstructured data.

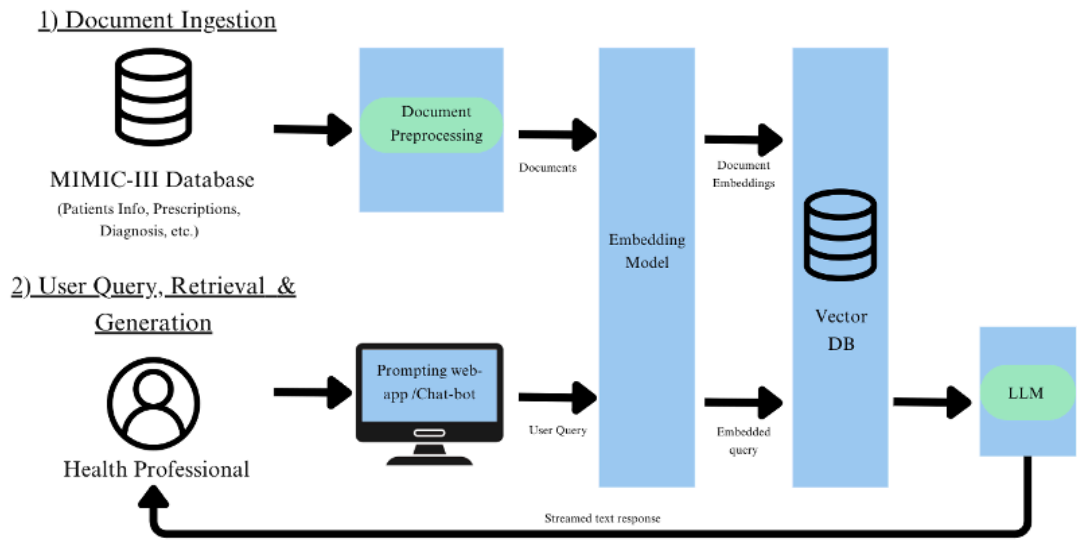

Figure 2. Retrieval Augmented Generation Pipeline

For both types of data, model and pipeline design choices were guided by the objective of evaluating a minimum viable prototype for LLM-based clinical data interaction. This approach prioritised architectural feasibility, reproducibility, and computational efficiency, allowing assessment of whether accurate structured querying and unstructured information extraction can be achieved under constrained but realistic deployment settings.

# Results

### a. Structured data

For the structured data experiments, we measured the accuracy of the models based on three categories of evaluation. The first was a boolean metric based on an exact match between the expected output and the model-generated output. The second one was an evaluation of the code generation correctness, based on the model's python code that interacts with the csv data. Thirdly was a human-annotated metric assessing the correctness of the model's interpretation, with three grades, allowing for partial matches when the underlying intent or reasoning was correctly captured despite minor differences in format or expression.

|  |  |  | Content correct of matches | | |
|---|---|---|---|---|---|
| Model | % of Exact Match Outputs | Code Correctness | Satisfactory | Partially Satisfactory | Not Satisfactory |
| llama 3 -8B Instruct | 3% | 60% | 7% | 43% | 50% |
| gpt_4o_mini | 50% | 73% | 40% | 33% | 27% |

Table 2. Structured data: NL to python results

### b. Unstructured data

For the RAG-based pipeline, extraction accuracy was evaluated using a combination of automated and human-centred metrics. Lexical overlap between the generated responses and the reference information was quantified using ROUGE scores, which provide a standardised measure of surface similarity. However, given the known limitations of ROUGE in capturing semantic equivalence, negation, and factual correctness; a complementary human-annotated boolean metric was also employed. This manual evaluation assessed whether each generated response was correct with respect to the underlying reference information, thereby providing a clinically meaningful measure of extraction accuracy that is not dependent on lexical overlap alone.

|  |  | Precision | | | Recall | | | F1 Score | | |
|---|---|---|---|---|---|---|---|---|---|---|
| Model | % of content correct of matches | R1 | R2 | RL | R1 | R2 | RL | R1 | R2 | RL |
| flan-t5-large | 76% | 0.67 | 0.51 | 0.67 | 0.29 | 0.21 | 0.27 | 0.35 | 0.26 | 0.35 |
| gpt-4o-mini | 78% | 0.61 | 0.44 | 0.56 | 0.51 | 0.39 | 0.48 | 0.50 | 0.37 | 0.49 |

Table 3. Unstructured data: RAG Results

# Conclusion

This study investigated the applicability of Large Language Models to two core Electronic Health Record data science tasks: querying structured clinical data through programmatic interfaces and extracting clinically relevant information from unstructured free-text records using a Retrieval-Augmented Generation pipeline. By evaluating both tasks within a unified experimental framework and on a shared clinical dataset, this work provides a comparative view of LLM capabilities across fundamentally different data modalities.

For structured data querying, results indicate that LLMs are capable of generating executable Pandas code that supports non-trivial analytical queries over clinical tables. However, performance varied substantially across models. While GPT-4o-mini achieved higher rates of exact-match outputs and code correctness, locally hosted models exhibited lower exact-match accuracy and a higher proportion of partially or unsatisfactory outputs. These findings highlight that, although LLMs can translate natural language questions into programmatic queries, reliability remains sensitive to model choice, and exact correctness cannot be assumed without validation.

For unstructured clinical text, the RAG-based pipeline demonstrated strong potential for supporting information extraction when retrieval mechanisms were used to ground model outputs. Human-annotated evaluations showed comparable levels of content correctness across models, while automated ROUGE-based metrics provided complementary insights into lexical overlap and coverage. Importantly, the observed discrepancies between ROUGE scores and human judgments reinforce that surface-level similarity metrics alone are insufficient for assessing factual correctness in clinical narratives, particularly in the presence of negation or partial summarisation.

Taken together, these results suggest that LLMs, when appropriately integrated with retrieval mechanisms and programmatic execution environments, can meaningfully support both structured analytics and unstructured information extraction in EHR contexts. At the same time, the findings underscore the necessity of task-aligned evaluation strategies that combine exact-match criteria, semantic similarity measures, and human judgment to ensure clinically meaningful assessment. Future work should explore scaling these approaches to broader clinical datasets, refining evaluation frameworks for domain-specific correctness, and integrating safeguards to support safe and reliable deployment in real-world clinical workflows.